\documentclass[conference]{IEEEtran}
\IEEEoverridecommandlockouts
\usepackage{cite}
\usepackage{amsmath,amssymb,amsfonts}
\usepackage{algorithmic}
\usepackage{graphicx}
\usepackage{textcomp}
\usepackage{xcolor}
\usepackage{comment}
\usepackage{tikz}
\usepackage{stfloats}
\usepackage{authblk}
\usetikzlibrary{arrows.meta, positioning}
\def\BibTeX{{\rm B\kern-.05em{\sc i\kern-.025em b}\kern-.08em
    T\kern-.1667em\lower.7ex\hbox{E}\kern-.125emX}}
    
\begin{document}

\title{Learning to Select Sparse Linear Solvers with Convolutional Neural Networks
\thanks{Department of Energy, National Nuclear Security Administration Award Number DE-NA0004266 and Defense Advanced Research Projects Agency (DARPA) Agreement No. HR00112490488.}
}

\begin{comment}
\author{\IEEEauthorblockN{Artemis Anna Pados}
\IEEEauthorblockA{\textit{Electrical Eng. and Computer Science} \\
\textit{Massachusetts Institute of Technology}\\
Cambridge, MA, USA \\
apados@mit.edu}
\and
\IEEEauthorblockN{Alan Edelman}
\IEEEauthorblockA{\textit{Electrical Eng. and Computer Science} \\
\textit{Massachusetts Institute of Technology}\\
Cambridge, MA, USA \\
edelman@mit.edu}
\and
\IEEEauthorblockN{Emmanuel Lujan}
\IEEEauthorblockA{\textit{Electrical Eng. and Computer Science} \\
\textit{Massachusetts Institute of Technology}\\
Cambridge, MA, USA \\
eljn@mit.edu}
\and
\IEEEauthorblockN{Daniel Pickard}
\IEEEauthorblockA{Aeronautics and Astronautics \\
\textit{Massachusetts Institute of Technology}\\
Cambridge, MA, USA \\
pickard@mit.edu}
\and
\IEEEauthorblockN{Felipe Tome}
\IEEEauthorblockA{\textcolor{red}{\textit{dept. name of organization (of Aff.)}} \\
\textcolor{red}{\textit{name of organization (of Aff.)}}\\
City, Country \\
felipe0@mit.edu}
\and
\IEEEauthorblockN{Christopher V. Rackauckas}
\IEEEauthorblockA{\textit{Electrical Eng. and Computer Science} \\
\textit{Massachusetts Institute of Technology}\\
Cambridge, MA, USA \\
crackauc@mit.edu}
}
\end{comment}

\author[1,4]{\fontsize{10}{12}\selectfont Artemis Pados}
\author[2,5]{Alan Edelman}
\author[2,6]{Emmanuel Lujan}
\author[3,7]{Daniel Pickard}
\author[2,8]{Felipe Tome}
\author[2,9]{Christopher Rackauckas}
\affil[1]{Department of Electrical Engineering and Computer Science, Massachusetts Institute of Technology, USA.}
\affil[2]{Computer Science \& Artificial Intelligence Laboratory, Massachusetts Institute of Technology, USA.}
\affil[3]{Department of Aeronautics and Astronautics, Massachusetts Institute of Technology, USA.}
\affil[4]{
\textit{apados@mit.edu}
\quad$^{5}$\textit{edelman@mit.edu}
\quad$^{6}$\textit{eljn@mit.edu}
\quad$^{7}$\textit{pickard@mit.edu}
\quad$^{8}$\textit{felipe0@mit.edu}
\quad$^{9}$\textit{crackauc@mit.edu}
}
\maketitle
\begin{abstract}
We investigate convolutional-neural-network (CNN) based automatic solver selection for sparse linear systems, a problem of practical importance where solver performance varies dramatically with matrix sparsity structure. A CNN is trained to regress log timing ratios between candidate solvers using spy plot images of matrix nonzero structure as input, achieving strong prediction accuracy on both held-out SuiteSparse matrices and real-world matrices from high-performance simulation codes.
\end{abstract}


\section{Introduction}

Sparse linear systems of the form $\mathbf{A}\mathbf{x} = \mathbf{b}$ where $\mathbf{A} \in \mathbb{R}^{n \times n}$ is a large sparse matrix arise throughout scientific computing, from finite element analysis and circuit simulation to fluid dynamics and structural mechanics applications. The efficiency of solving such systems is critical to the scientific application and the choice of direct solver can have a dramatic impact. For the same matrix, different solvers can vary in wall-clock time by orders of magnitude, ranging from milliseconds to several seconds ~\cite{lujan2025hpec}.
 
Julia~\cite{julia} has emerged as a leading language for high-performance scientific computing, combining the productivity of high-level languages with the performance of compiled code. Its \textit{LinearSolve.jl}~\cite{linearsolve} package provides a unified interface to multiple high-performance direct sparse solvers. Despite this flexibility, current software, including MATLAB, Python/SciPy, and Julia, relies on fixed default backends regardless of the structural properties of the input matrix. This means that users who do not manually benchmark solvers on their specific matrices are likely leaving significant performance gains on the table.
 
In this paper, we propose the use of convolutional neural networks (CNNs) as a data-driven approach to automatic solver selection, trained to regress log timing ratios collected across a large corpus of sparse matrices from the SuiteSparse Matrix Collection~\cite{suitesparse}. Further, we release \textit{NeuralLinearSolve.jl}\footnote{https://github.com/SciML/NeuralLinearSolve.jl}, an open-source Julia package integrating the trained model bundled with frozen weights into the SciML ecosystem\footnote{https://sciml.ai/} and the broader \textsc{SmartSolve} effort~\cite{lujan2025} at the MIT Julia Lab, exposing solver recommendations via a single function call with no benchmarking or domain expertise required.

\section{Related Work}
\label{background}

Direct sparse linear solvers have been extensively studied, with different methods optimized for different matrix structures, yielding no single solver dominating across problem types~\cite{lujan2025hpec, ferrari2024}. Davis~\cite{davis2006} provides a comprehensive overview of solvers including those that we consider: UMFPACK implements multifrontal LU and is Julia's default, KLU uses a left-looking sparse LU optimized for nearly-triangular matrices, and MKL Pardiso is Intel's parallel supernodal direct solver. For symmetric positive definite matrices, CHOLMOD~\cite{cholmod} selects between solvers using a reliable hand-engineered heuristic. For general sparse matrices, however, no such heuristic exists, motivating a learning-based approach. Recent work in this direction has explored LLM-driven heuristic generation~\cite{shah2025}. Yamada et al.~\cite{yamada2018} propose a similar CNN-based approach but for classification of preconditioners in iterative solvers, using full-color RGB feature images.

\section{Our Methodology}

\subsection{Data Collection}
 
We benchmark the three solvers mentioned in Section~\ref{background} across approximately 1500 square, real-valued sparse matrices from the SuiteSparse Matrix Collection~\cite{suitesparse}. Wall-clock solve times are measured using a minimum-over-trials strategy to reduce sensitivity to OS scheduling noise and transient system load. The fastest solver per matrix serves as the ground-truth label. The intentional imbalance of the resulting dataset is addressed during training via class-weighted loss.

\subsection{Spy Plot Generation and CNN}
 
Each matrix is converted to a $64 \times 64$ binary spy plot image, where each pixel indicates the presence of at least one nonzero in the corresponding matrix block. Matrices larger than $64 \times 64$ are downsampled via block aggregation, whereas smaller matrices are upsampled via nearest-neighbour interpolation. The global sparsity pattern of the original matrix captures the structural properties that determine solver suitability: near-triangular structure favors KLU's reordering strategies, dense or irregular patterns favor UMFPACK's multifrontal approach, and larger structured problems favor Pardiso's parallelism. These coarse-grained features are well-preserved at $64 \times 64$ resolution.
 
% \subsection{CNN Architecture and Training}

\begin{comment}
    The CNN is trained on pairs of spy plot images and their corresponding timing labels, learning to map visual sparsity structure directly to solver performance. The architecture consists of three convolutional blocks, each comprising a $3\times3$ convolution with ReLU activation and $2\times2$ max pooling, with filter counts increasing from 1 to 16, 32, and 64. The resulting $4096$-dimensional feature vector is passed through three fully connected layers ($4096\to256\to64\to2$) with ReLU activations and dropout, yielding two scalar outputs.
\end{comment}

The CNN is trained on pairs of spy plot images and their corresponding timing labels, learning to map visual sparsity structure directly to solver performance. The architecture consists of three convolutional blocks and three fully connected layers with ReLU activations, max pooling, and dropout, yielding two scalar outputs.

Rather than treating solver selection as a three-class classification problem, we formulate it as regression on log timing ratios. KLU serves as the denominator baseline since it is the fastest solver for the majority of matrices in our training dataset, making it the most stable reference point. Log ratios are used in place of raw ratios to symmetrize the target distribution and reduce sensitivity to outliers. The regression targets are
% \begin{equation}
$    \hat{y} = \left[ \log\frac{t_{\text{UMFPACK}}}{t_{\text{KLU}}},\ \log\frac{t_{\text{Pardiso}}}{t_{\text{KLU}}} \right]$
% \end{equation}
and the predicted solver is $\hat{s} = \operatorname{argmin}\left(\hat{y}_1,\ 0,\ \hat{y}_2\right)$, where 0 is the implicit KLU score. This formulation encourages the model to learn the magnitude of performance differences rather than just which solver wins, yielding more robust predictions near decision boundaries. The model is trained using MSE loss, the Adam optimizer, and an 80/20 train/test split.
 
\subsection{Julia Package}
 
The trained weights are bundled as a BSON artifact within \textit{NeuralLinearSolve.jl}. The package exposes two functions: predict\_solver(A) and predict\_solver\_scores(A), returning the recommended solver and relative speed scores. Both accept a matrix as input and generate the spy plot internally.

\section{Experimental Results}

% \subsection{Training and Validation}
 
\begin{figure}[htbp]
\centerline{\includegraphics[width=0.93\columnwidth]{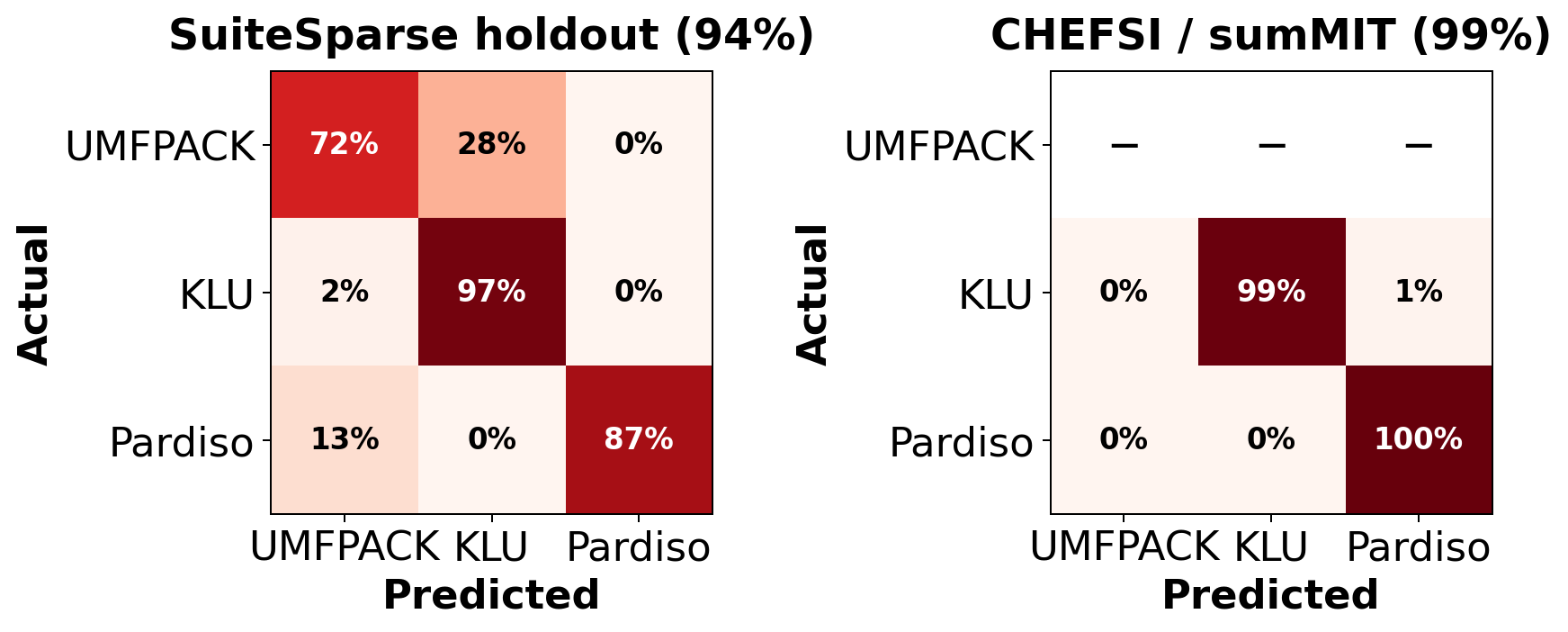}}
\caption{Normalized confusion matrices for SuiteSparse holdout (left) and CHEFSI/$\Sigma$MIT evaluation (right).}
\label{confusion}
\end{figure}

 On the held-out test set (20\% of the corpus),  the model achieves \textbf{94\%} accuracy in predicting the fastest solver (Fig. \ref{confusion}). The improvement (over 84\% that would have been achieved if we fixed KLU as the solver) is concentrated in the minority classes where the performance gap between solvers is largest.
 
% \subsection{Evaluation on CHEFSI and $\Sigma$MIT Stiffness Matrices}
 
To evaluate generalization beyond the SuiteSparse training distribution, we test \textit{NeuralLinearSolve.jl} on a corpus of matrices harvested from real high-performance scientific computing applications, namely $\Sigma$MIT\footnote{http://summit.mit.edu} and CHEFSI\footnote{https://chefsi.mit.edu/}. On this corpus, \textit{NeuralLinearSolve.jl} achieves \textbf{99\%} prediction accuracy (Fig. \ref{confusion}). 
 
\section{Conclusion and Future Work}

We have presented \textit{NeuralLinearSolve.jl}, a Julia package that uses a pretrained CNN to automatically predict the fastest direct sparse linear solver for a given matrix from its spy plot image, achieving strong accuracy on both SuiteSparse held-out matrices and real-world stiffness matrices from CHEFSI and $\Sigma$MIT simulations.
 
Several directions are planned for future work. First, we aim to expand the solver set beyond the current three backends to include additional solvers available in \textit{LinearSolve.jl}. More ambitiously, we plan to extend the framework to the joint selection of solver and preconditioner (PC) combinations using a dedicated PETSc benchmarking engine\footnote{https://github.com/Smart-Solve-AI/SmartDiscovery}. This work would constitute a strict superset of \textit{NeuralLinearSolve.jl}. Ultimately, we aim to conduct interpretability analyses and integrate the learned selection model directly into \textit{LinearSolve.jl} and $\Sigma$MIT, enabling near-optimal solver dispatch for practitioners with no domain expertise required.

\section*{Acknowledgment}

This material is based upon work supported by the Department of Energy, National Nuclear Security Administration under Award Number DE-NA0004266 and by the Defense Advanced Research Projects Agency (DARPA) under Agreement No. HR00112490488. Claude Sonnet 4.6 (Anthropic) was used to revise full author-written drafts for clarity, coherence, and correct typographical errors.

\end{document}